\documentclass{article}
\usepackage{spconf,amsmath,graphicx}

\title{Painting Style-Aware Manga Colorization\\ Based on Generative Adversarial Networks}
\name{Yugo Shimizu$^{\star}$ \qquad Ryosuke Furuta$^{\dag}$ \qquad Delong Ouyang$^{\star}$ }
\secondlinename{Yukinobu Taniguchi$^{\star}$ \qquad Ryota Hinami $^{\ddag}$ \qquad Shonosuke Ishiwatari$^{\ddag}$ }
\address{$^{\star}$ Tokyo University of Science \qquad $^{\dag}$ The University of Tokyo\qquad $^{\ddag}$ Mantra Inc.}

\begin{document}
%
\maketitle
\begin{abstract}
Japanese comics (called manga) are traditionally created in monochrome format. In recent years, in addition to monochrome comics, full color comics, a more attractive medium, have appeared. Unfortunately, color comics require manual colorization, which incurs high labor costs. Although automatic colorization methods have been recently proposed, most of them are designed for illustrations, not for comics. Unlike illustrations, since comics are composed of many consecutive images, the painting style must be consistent. 

To realize consistent colorization, we propose here a semi-automatic colorization method based on generative adversarial networks (GAN); the method learns the painting style of a specific comic from small amount of training data. The proposed method takes a pair of a screen tone image and a flat colored image as input, and outputs a colorized image. Experiments show that the proposed method achieves better performance than the existing alternatives.
\end{abstract}

\section{Introduction}
\label{sec:intro}
Japanese comics, also known as manga, are read not only in Japan, but around the world, and their popularity continues to grow. Although they have been traditionally drawn and sold in monochrome format with screen tones, full color comics are now being sold in because they attract a wider range of customers. However, the colorization process is time-consuming and costly, which is a huge problem. 

In order to solve the above problem, many automatic colorization methods based on machine learning have been proposed for illustrations. However, their results are not pleasant enough for practical use. In addition, they have two problems: (i) none of them pay attention to consistency in painting style between images on different pages, and (ii) most of them require a large amount of training data. Failure of a colorized comic to maintain consistency in painting style across the pages degrades the reader experience and enjoyment. A large amounts of training data are difficult to obtain in many cases due to copyright issues.

In this paper, we propose a painting style-aware colorization method based on GANs. The proposed method takes a pair of a screen tone image and a flat colored image as input, and outputs a colored image. Because the shadow and color information are given by the screen tone and flat colored image, respectively, the generator does not need to learn or infer those for the objects in the input image. Therefore, the proposed method (i) focuses on and learns only the painting style in the training data (ii) and can be trained with small amounts of data. We use flat colored images as input because they are easy to create and contain a wealth of information as hints.

We conducted experiments on the comics in the Manga109 dataset~\cite{manga109_1}~\cite{manga109_2} and commercially available comics. The experimental results show that the proposed method can efficiently learn the painting style from a small amount of data and achieve better performance than other methods.


\section{Related Works}
\begin{figure}[t]
  \centerline{\includegraphics[width=8cm]{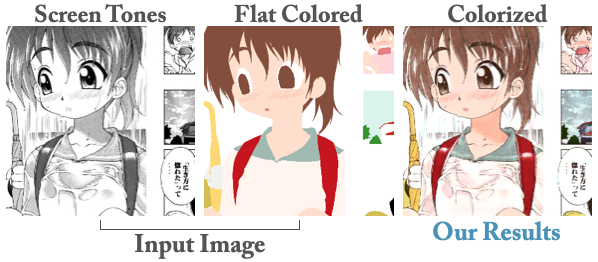}}
\caption{Input and output images of our method.}
\label{fig:sample}
\end{figure}
\label{Relevant}
Various colorization methods have been already proposed, for not only illustrations or manga but also natural images, as is well summarized in~\cite{summary}. This paper focuses on works that relevant to illustrations and manga.

\subsection{Illustration colorization}
Existing colorization methods are categorized into two classes: those based on hand-crafted algorithms such as~\cite{tj1}~\cite{tj2}, and~\cite{tj3} and machine learning. We focus on the latter because the proposed method employs machine learning.

In order to create the dataset, in general, it is necessary to prepare a large number of colored images and their corresponding line drawings. Liu et al.~\cite{autopainter} create their line drawings by applying a XDoG filter to the colored images. 
However, their method requires manual adjustments, and the painting styles output are not close to the ground truth images although their method can colorize specific regions with the correct color.

Zhang et al.~\cite{2stage} proposed a method based on Pix2Pix, where user hints are added in addition to line drawings as their input. They use the danbooru dataset~\cite{danbooru} as their training data. Although their method can successfully learn the colorization for illustrations, directly adopting their method to comics has several drawbacks. In the colorization of comics, each comic has a different painting style. However, because model training uses a large number of colored images from different artists, it cannot learn the painting style of a particular comic. In addition, unlike our method, which uses screen tone images as input, the position of shadows cannot be specified in their method.

Recently, Ren et al.~\cite{22} proposed a two-stage sketch colorization method, and Akita et al.~\cite{23} proposed a colorization method that uses a reference image. Zou et al.~\cite{24} proposed a language-based colorization method, where users can control colorization by using text descriptions as input. Different from the above methods, as discussed in Section 1, we propose a style-aware colorization method for manga by taking a pair of a screen tone image and a flat colored image as input. 

Alternatives include some attempts to base illustration on machine learning~\cite{25},~\cite{26},~\cite{27}, and~\cite{28}. Different from ours, their target is learning for lighting effects or shadowing, not colorization.

\subsection{Manga colorization}
Hensman and Aizawa~\cite{cgan} proposed an automatic manga colorization method. In this method, the conditional GAN (cGAN) is trained online using a single reference image. This method has the advantage that a large amount of training data is not needed. However, it is sometimes difficult to obtain reference images similar to the target image (e.g., some characters appear only a few times in a comic).

Xie et al.'s method~\cite{VAE} enables the bidirectional translation between screen tone images and colorized images. However, this method requires a large amount of training data, which is difficult to obtain in most cases.

Silva et al.~\cite{ICIP} proposed a hint-based semi-automatic manga colorization method that uses cGAN as its model. The difference between ManGAN and our method is that ManGAN takes a pair of line-art and color-hint images as input, whereas ours takes a pair of screentones and flat-colored images. Moreover, the network architectures are different.
 
\begin{figure}[t]
  \centerline{\includegraphics[width=6cm]{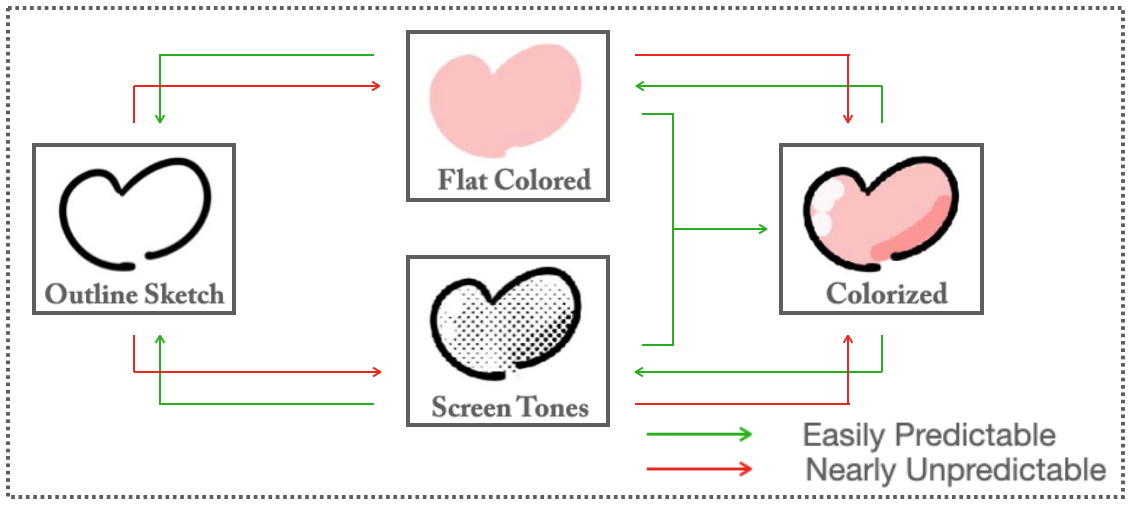}}
\caption{Name and Relationships between manga states.}
\label{fig:relation}
\end{figure}
\begin{figure}[t]
  \centerline{\includegraphics[width=7cm]{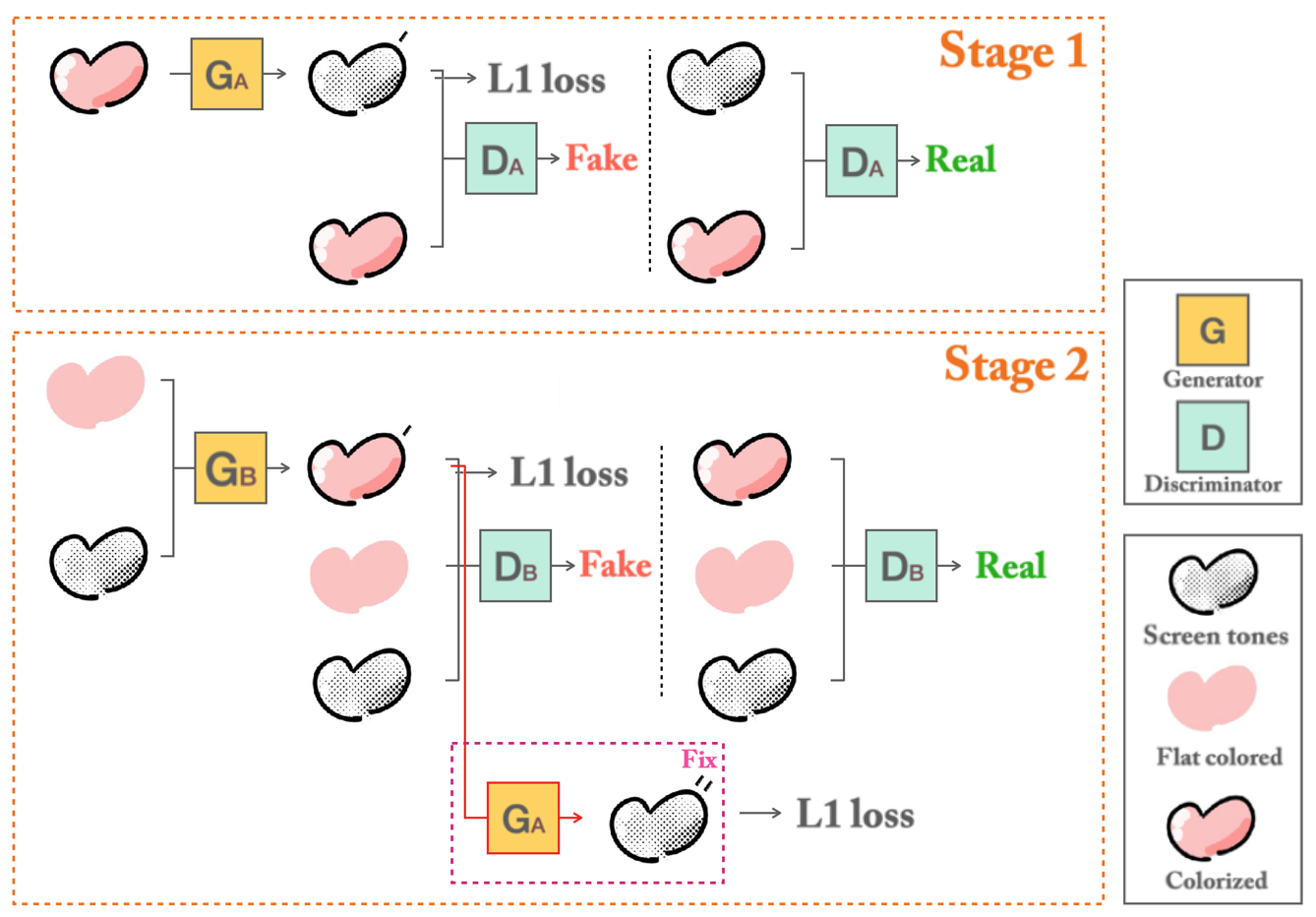}}
\caption{Structure of our proposed method.}
\label{fig:promethod}
\end{figure}
\section{Proposed method}
\subsection{Observations}
Before detailing the proposed method, we discuss the states of manga images and their relationships. Fig.~\ref{fig:relation} shows the name of each state and the relationships between them. The green arrows indicate the conversions between the states that are easy to predict, while the red arrows indicate the nearly unpredictable conversions (due to insufficient information). Because an outline sketch does not contain shading or color information, it is difficult to determine the direction of the light and the appropriate color uniquely (i.e., it is difficult to predict the colorized image from the outline sketch as shown in Fig.~\ref{fig:relation}). Zhang et al.~\cite{2stage} solved this problem by learning from a large amount of training data. However, such training data is not always available. This is especially true when we want to learn the painting style of a particular colorist because it is difficult to create a large-scale dataset that contains only the artworks drawn by the particular person. Similarly, due to lack of information, to predict the colored image from either flat colored or screen tone image is not ideal. As shown in Fig.~\ref{fig:relation}, the direction of the light and the appropriate color can be uniquely determined by using both flat colored and screen tone images, which makes it possible to predict the colorized image. Therefore, in this paper, we propose a method that predicts the colorized image by using both flat colored and screen tone images as input. 

Because the screen tone images represent the monochrome manga itself, it incurs no additional costs to obtain the screen tone images. The user only needs to prepare the flat colored images. Compared to the actual coloring process, preparing flat color images requires no special skills and much less time and effort, and thus anyone (not just colorists) can perform the task.
\subsection{Framework}
The training process is shown in Fig.~\ref{fig:promethod}.

The proposed method has two generators. One converts a colorized image into a screen tone image, while the other generates a colorized image from a pair of flat colored and screen tone images. These two generators are trained separately. The task of converting a colorized image to a screen tone image is called ``stage 1'', and the generation of a colorized image from a pair of flat colored and screen tone images is called ``stage 2''.

The training data consists of the triplets of colorized, screen tones, and flat colored images ($x$, $y$, $z$).
First, in ``stage 1'', generator $G_A$ learns how to generate screen tone images from colorized images. This process removes color information from colorized images and predicts the position and patterns of the corresponding screen tones. As shown in Fig.~\ref{fig:relation}, colorized images contain sufficient information to predict screen tone images. The training procedure follows that of Pix2Pix~\cite{pix2pix}.
\subsubsection{training stage 1}
We adopted the UNet architecture for generator $G_A$. Let x denote a colorized image and y denote a screen tone image. The discriminative loss of $G_A$ is expressed as:
\begin{equation}
\begin{split}
	\mathcal{L}_{G_A}(G_A,D_A)=E_{x,y}[\log D_A(x,y)]+\\E_{x}[\log (1-D_A(x,G_A(x)))],  \label{1}
\end{split}
\end{equation}
where generator $G_A$ learns how to fool the discriminator $D_A$. Conversely, discriminator $D_A$ learns to discriminate between fake and real tuples.

In addition to the discriminative loss in Eq.~\ref{1}, we use L1 loss between the ground truth screen tone image y and the generated image $G_A(x)$. Because pix2pix uses L1 loss in order to suppress blurring, we used the same L1 loss term in the proposed method.:
\begin{equation}
	\mathcal{L}_{L_1}(G_A)=E_{x,y}[\| y-G_A(x) \|_1 ].  \label{2}
\end{equation}
Our final objective of $G_A$ is:
\begin{equation}
	G_A^*= \arg \mathop{\min}_{G_A}\mathop{\max}_{D_A}
	\mathcal{L}_{G_A}(G_A,D_A)+
	\lambda_1\mathcal{L}_{L_1}(G_A). \label{3}
\end{equation}

\subsubsection{training stage 2}

After completion of ``stage 1'' training, we move on to ``stage 2''. In ``stage 2'', the input is a pair of flat colored and screen tone images. The generator $G_B$ learns how to generate colorized images from flat colored and screen tone images. The generator model is an extension of ``U-Net''~\cite{unet}. In order to gain one output from two inputs, the model has a two stream structure.

Let $x$, $y$, and $z$ be a colorized image, a screen tone image, and a flat colored image, respectively. The discriminative loss of $G_B$ is expressed as:
\begin{equation}
\begin{split}
	\mathcal{L}_{G_B}(G_B,D_B)=E_{x,y,z}[\log D_B(y,z,x)]+\\E_{y,z}[\log (1-D_B(y,z,G_B(y,z)))],  \label{4}
\end{split}
\end{equation}
where the generator $G_B$ learns to fool the discriminator $D_B$. In contrast, the discriminator $D_B$ learns to classify between the fake and real triplets. We also use $L_1$ loss to increase the quality of the output:
\begin{equation}
	\mathcal{L}_{L_1}(G_B)=E_{x,y,z}[\| x-G_B(y,z) \|_1 ],  \label{5}
\end{equation}

Furthermore, in order to retain cycle consistency\cite{cyclegan}, the colorized image generated by $G_B$ is input to the fixed $G_A$. The $L_1$ distance between the fake screen tone image from the fixed $G_A$ and the ground truth screen tone image is calculated (the second term in Eq.~\ref{6}). This idea,  based on~\cite{discogan}~\cite{cyclegan}, contributes to increasing the quality of the output. Our final objective in stage 2 is:
\begin{equation}
\begin{split}
	G_B^*= \arg \mathop{\min}_{G_B}\mathop{\max}_{D_B}
	\mathcal{L}_{G_B}(G_B,D_B)+
	\lambda_2\mathcal{L}_{L_1}(G_A) +
	\\\lambda_3\mathcal{L}_{L_1}(G_B). \label{6}
\end{split}
\end{equation}
\subsubsection{Testing}
At the time of inference, we can obtain the colorized image by inputting a pair of screen tone image and its corresponding flat colored image into generator $G_B$.

\subsection{Discussions}
Because flat colored images and screen tone images contain color and shading information, respectively, there is no need to predict the color and shading of the objects in the input. The proposed model can focus on and learn only the painting style in the training data. Therefore, our method can be trained with only a small amount of data and can learn the painting style of a particular colorist.

\section{Experiments}
\subsection{Dataset}
\textbf{Manga109}: Manga109~\cite{manga109_1}~\cite{manga109_2} is a dataset that consists of 109 volumes of Japanese manga drawn by professional manga artists. We randomly selected 10 pages from manga ``Nekodama $\copyright$Ebifly''. We used five pages as the training data, and the remaining pages as the test data. Images in the training data were colorized and flat colored manually by a proficient colorist using CLIP STUDIO PAINT EX.\\
\textbf{Commercial dataset A}: We used ``Sgt. Frog'', a commercially available comic. Because the colorized data is available only for the cover pages, we used them as ground truth. We generated screen tone images from those cover pages by passing them through the LT filter. The flat coloring process was done manually.  We used 30 cover pages from ``Sgt. Frog'', and used the first ten pages as the training data.\\
\textbf{Commercial dataset B}: We also used commercial comics available in both monochrome and colored versions. In this dataset, because we have both colorized and screen tone images, we prepared only the flat colored images manually. We randomly selected 10 pages from ``Gotobun no Hanayome'', and used the first two pages as the training data.

We report only the quantitative results in the commercial datasets A/B due to copyright issues.

\subsection{Compared methods}
We compare the proposed method with the following four methods in terms of performance. \\
\textbf{(i).Pix2Pix (Screen Tones)}: We trained Pix2Pix, which takes a screen tone image as input and outputs a colored image.\\
\textbf{(ii).Pix2Pix (Flat Colored)}: We trained another Pix2Pix model. Different from (i), this model takes a flat colored image as input.\\
\textbf{(iii).cGAN-Based Manga Colorization}: This method requires a reference image. For details, see ~\cite{cgan}.\\
\textbf{(iv).Two-Stage Sketch Colorization}~\cite{2stage}: We used a third party implementation of~\cite{2stage} \footnote{https://github.com/adamz799/Paints}. This model was trained on the danbooru dataset~\cite{danbooru}.

(i) and (ii) were trained on the same training sets as the proposed method.
\begin{figure}[t]
  \centerline{\includegraphics[width=8cm]{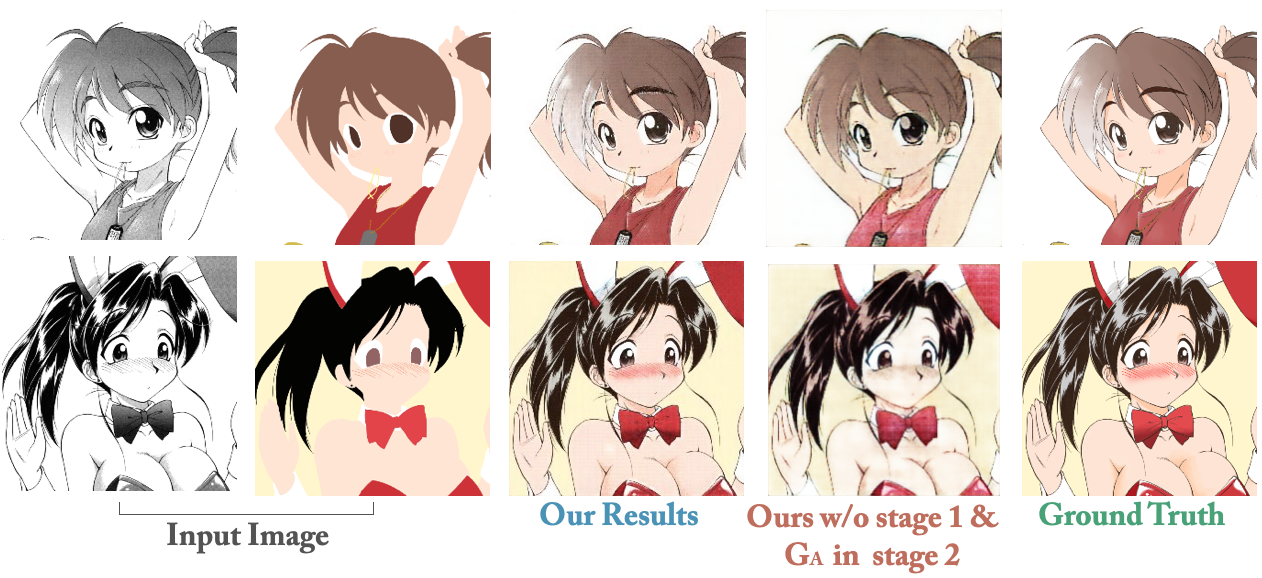}}
\caption{Ablation study of the proposed method.}
\label{fig:res1}
\end{figure}
\begin{figure}[t!]
 \centerline{\includegraphics[width=7
cm]{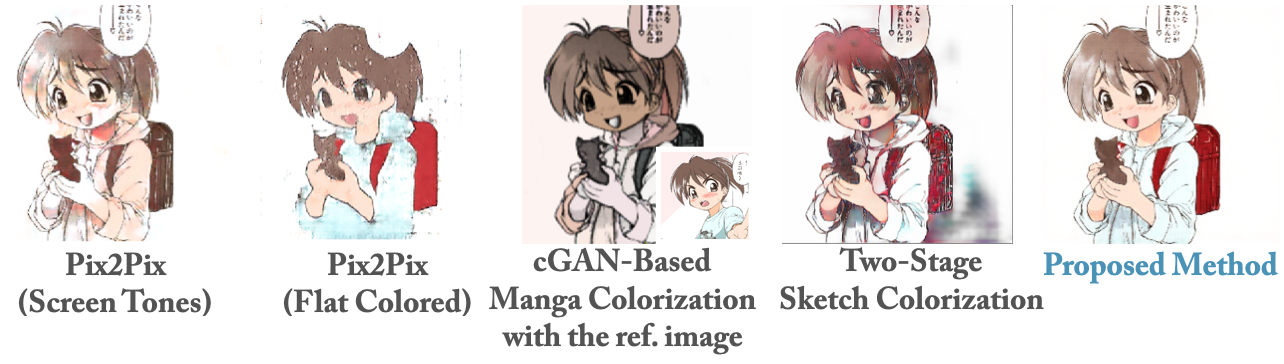}}
\caption{Qualitative comparisons.}
\label{fig:res0}
\end{figure}
\begin{table}[t]
\centering
\caption{PSNR on Commercial datasets A and B.}
  \begin{tabular}{|c|c|c|c|c|c|} \hline
 & & (i)~\cite{pix2pix} & (ii)~\cite{pix2pix} & (iv)~\cite{2stage} & Ours \\ \hline \hline
   & Ave. & 12.99 & 15.22 & 15.19 & \bf{26.71} \\
dataset A  & Max & 15.27 & 19.28 & 17.46 & \bf{27.38}\\
   & Min & 8.47 & 9.86 & 13.68 & \bf{25.71} \\\hline
   & Ave. & 17.95 & 16.17 & 13.26 & \bf{24.47} \\
dataset B  & Max & 20.64 & 18.47 & 16.66 & \bf{27.02}\\
   & Min & 14.39 & 14.28 & 11.00 & \bf{21.73} \\\hline
  \end{tabular}
  \label{tab:datasetA}
\end{table}
\subsection{Results}
Fig.~\ref{fig:res1} shows the ablation results of the proposed method by comparing the results w/ and w/o stage 1 and $G_A$ in stage 2 (see Fig.~\ref{fig:promethod}). We observe that by incorporating them, the proposed method successfully learned the painting style of the ground truth image created by the colorist.

Fig.~\ref{fig:res0} shows qualitative comparisons of the proposed method and the other methods. Compared to cGAN-based method~\cite{cgan} and Two-stage method~\cite{2stage}, the proposed method yielded much more pleasing results. We observe that Pix2Pix (Screen Tones) and Pix2Pix (Flat Colored) failed to predict the colors and shadow, respectively, due to the small amount of training data, and the insufficient information in their inputs. In contrast, the proposed successfully learned the painting style by taking both screen tones and flat colored images as input.

Table~\ref{tab:datasetA} shows the comparisons of PSNR on the commercial datasets A and B. We observe that our method achieved significantly better results than the alternative methods.


\section{Conclusion}
In this paper, we proposed a style-aware colorization method for manga. The proposed method can learn unique painting styles from a small amount of training data by taking a pair of screen tones and flat colored images as input. The outputs of our method exhibit very high visual quality because the generator efficiently learns the painting style from a particular colorist. Because the flat coloring process does not require any painting skill, anyone can perform this task, which means our method can replace the traditional cumbersome steps.

As a practical workflow, the colorist only needs to colorize the first few pages manually. We use these images as the training data for the proposed method. The method automatically colorizes the remaining pages by preparing the corresponding flat colored images and inputting them to trained generator $G_B$. The proposed workflow makes it possible to reduce time and money overheads.

In future work, we will tackle automatic generation of flat colored images, which makes our method more practical.





\label{sec:ref}

\bibliographystyle{IEEEbib}

\begin{figure*}[t]
  \centerline{\includegraphics[width=15cm]{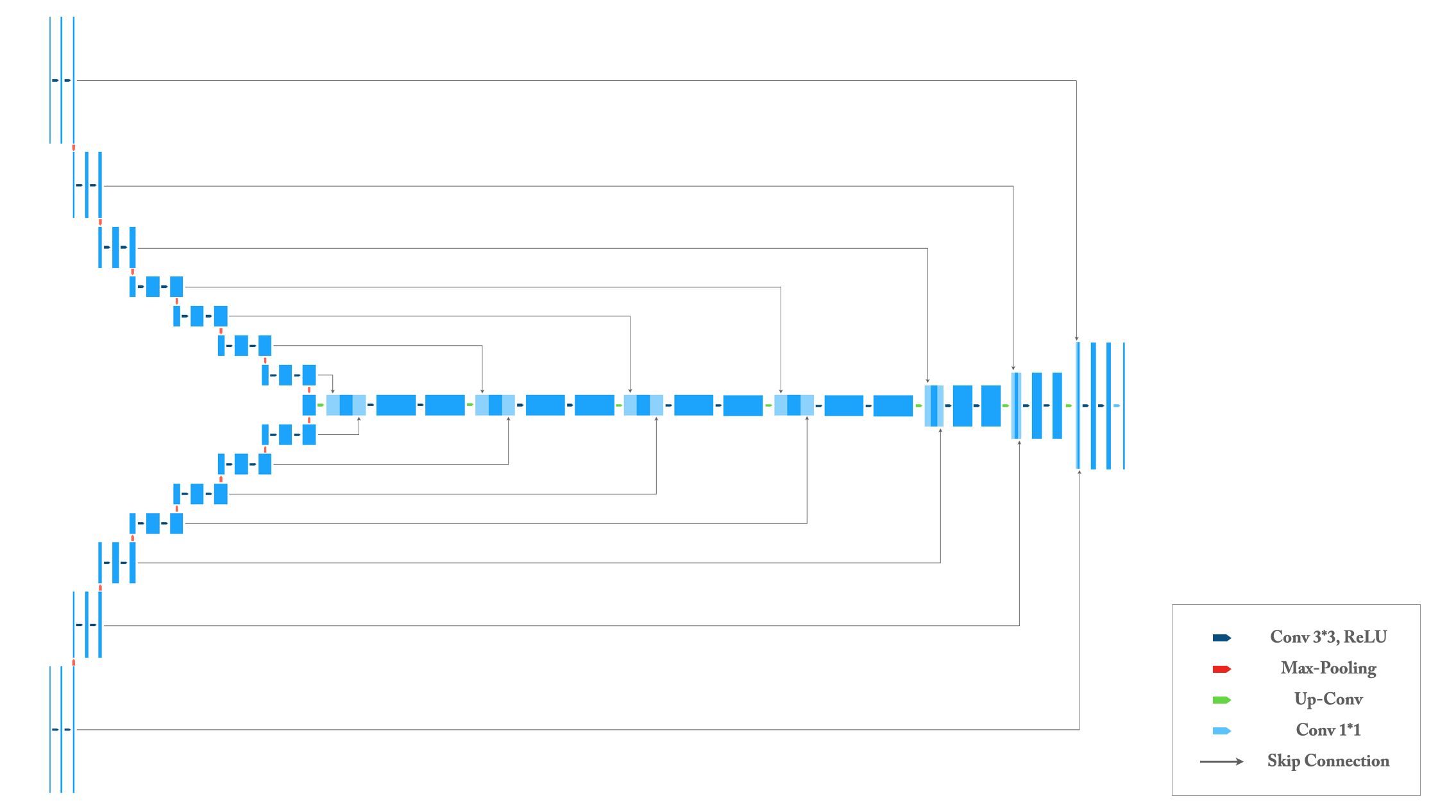}}
\caption{Network structure of generator $G_B$.}
\label{fig:model}
\end{figure*}
\begin{figure}[t!]
  \centerline{\includegraphics[width=8cm]{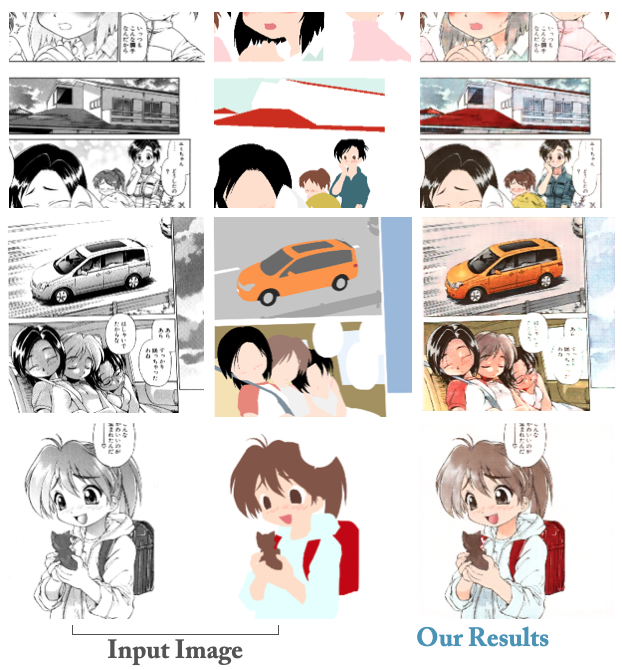}}
\caption{Images from Manga109 colorized by our method.}
\label{fig:res2}
\end{figure}
\begin{figure*}[t!]
  \centerline{\includegraphics[width=18cm]{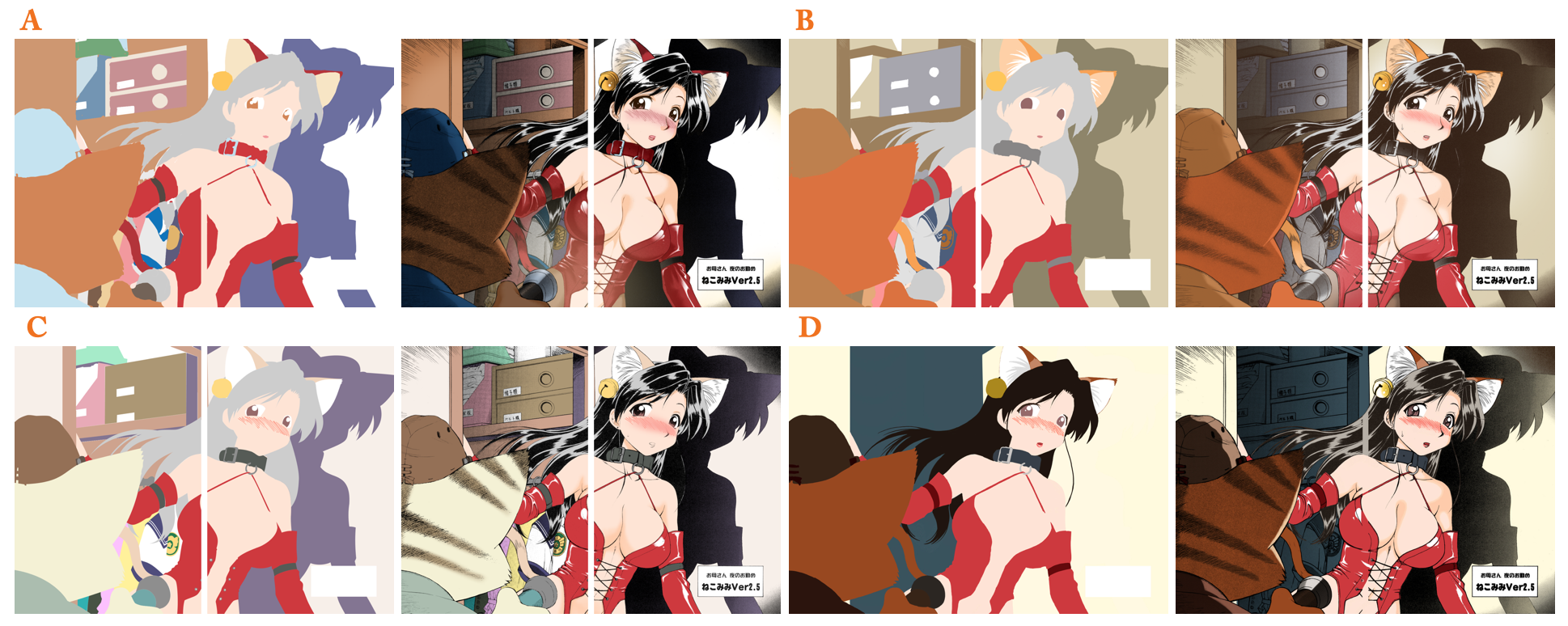}}
\caption{Images colorized by four professional illustrators.}
\label{fig:res3}
\end{figure*}
\begin{table}[t]
\centering
\caption{Time required for manually preparing flat colored images and colorized images.}
  \begin{tabular}{|c|c|c|} \hline
 & Flat colored $(min)$ & Colorized $(min)$ \\ \hline \hline
Illustrator A & \bf{73}   &   148 \\ 
Illustrator B & \bf{80}   &   172 \\
Illustrator C & \bf{160}  &   320\\
Illustrator D & \bf{75}   &   180 \\\hline \hline
Average & \bf{97} & 205\\\hline

  \end{tabular}
  \label{tab:mangaka}
\end{table}
\section{Supplemental Material}
\subsection{Implementation details}
The generator model is an extension of ``U-Net''~\cite{unet} as shown in Fig.~\ref{fig:model}. In order to gain one output from two inputs, the model has a two stream structure. Each input image is processed by different encoders, and the feature maps are extracted by convolution. Then, the feature maps from the two encoders are concatenated and convolved. The resulting feature map is used as the input of the decoder. Similar to U-Net, this model also has skip connections between some corresponding layers in the encoder and the decoder.

We implemented the proposed method using the PyTorch library~\cite{pytorch}. We used Adam optimizer and set the learning rate to 0.001. The number of training epochs was 150, 100, and 100 for the Manga109 dataset, commercial dataset A, and B, respectively. Each page was resized to $1024\times 724$ and $640\times 453$, and then randomly cropped to $256\times256$ during the training. We set $(\lambda_1,\lambda_2,\lambda_3)$ as $(100,50,50)$.

We use the same notation as ~\cite{pix2pix}. Ck denotes a Convolution-BatchNorm-ReLU layer with k filters, and CDk denotes a Convolution-BatchNorm-Dropout-ReLU layer with a dropout rate of $0.5$. The encoder architecture of generator $G_A$ is C64-C128-C256-C512-C512-C512-C512, and their decoder architecture is CD512-CD512-CD512-CD512-CD256-CD128-CD64. All convolutions are $3*3$ spatial filters applied with stride $1$. Generator $G_B$ has the same architecture as $G_A$ except that it has two stream encoders. Similar to~\cite{pix2pix}, there are some exceptions with regard to BatchNorm and ReLU in some layers. The first C64 layer does not have BatchNorm, and all ReLUs in the encoder are leaky with slope 0.2 while ReLUs in the decoder are not leaky.

The discriminator architecture is C64-C128-C128. Similar to the original paper of Pix2Pix~\cite{pix2pix}, after the last layer, a convolution is applied to map to a 1-dimentional output, followed by a Sigmoid function. The first C64 layer does not have BatchNorm, and all ReLUs are leaky with slope 0.2.

\subsection{Additional results}
\subsubsection{Qualitative results}
Fig.~\ref{fig:res2} shows the input images and the output results of the proposed method on the Manga109 dataset. We observe that the proposed method yielded high quality results.
\subsubsection{Comparison of time for manual colorization}
Although the proposed method requires flat colored images as input, they are easier to create than colorized images as mentioned in Sec. 1. In order to ensure this premise, we conducted the user study to measure the time required for manually preparing flat colored images and colorized ones, respectively. We asked four professional illustrators to manually create them from a screen tone image. We used a page randomly chosen from ``Nekodama $\copyright$Ebifly''. Table ~\ref{tab:mangaka} and Fig.~\ref{fig:res3} show the time and result images, respectively. We observe that flat coloring process can save $53\%$ of time on average.

\end{document}